\documentclass[journal]{IEEEtran}
\IEEEoverridecommandlockouts
\usepackage{cite}
\usepackage{amsmath,amssymb,amsfonts}
\usepackage{algorithm}
\usepackage{graphicx}
\usepackage{textcomp}
\usepackage{xcolor}
\usepackage{algpseudocode}
\usepackage{verbatim} 
\usepackage{comment}
\usepackage{xcolor}
\usepackage{svg}
\usepackage{amssymb}
\usepackage{xspace}
\usepackage[numbers]{natbib}
\usepackage{multirow}
\usepackage{tabularx}
\usepackage{booktabs}
\usepackage{subcaption}
\usepackage[font=footnotesize]{caption}
\usepackage{graphicx}
\usepackage{geometry}
\begin{document}

\title{FedGreen: Carbon-aware Federated Learning with Model Size Adaptation}

\author{
\IEEEauthorblockN{Ali Abbasi\IEEEauthorrefmark{1}, Fan Dong\IEEEauthorrefmark{1},  Xin Wang\IEEEauthorrefmark{2}, Henry Leung\IEEEauthorrefmark{1}, 
Jiayu Zhou\IEEEauthorrefmark{3}, Steve Drew\IEEEauthorrefmark{1}}

\IEEEauthorblockA{
\IEEEauthorrefmark{1}Department of Electrical and Software Engineering, University of Calgary, Calgary, AB, Canada \\
\IEEEauthorrefmark{2}Department of Geomatics Engineering, University of Calgary, Calgary, AB, Canada \\
\IEEEauthorrefmark{3}Department of Computer Science and Engineering, Michigan State University, East Lansing, MI, USA \\
\{ali.abbasi1, fan.dong, xcwang, leungh\}@ucalgary.ca, zhou@cse.msu.edu, steve.drew@ucalgary.ca}
}

\maketitle

\thispagestyle{plain}
\pagestyle{plain}

\begin{abstract}
Federated learning (FL) provides a promising collaborative framework to build a model from distributed clients, and this work investigates the carbon emission of the FL process.  
Cloud and edge servers hosting FL clients may exhibit diverse carbon footprints influenced by their geographical locations with varying power sources, offering opportunities to reduce carbon emissions by training local models with adaptive computations and communications. 
In this paper, we propose FedGreen, a carbon-aware FL approach to efficiently train models by adopting adaptive model sizes shared with clients based on their carbon profiles and locations using ordered dropout as a model compression technique. 
We theoretically analyze the trade-offs between the produced carbon emissions and the convergence accuracy, considering the carbon intensity discrepancy across countries to choose the parameters optimally. 
Empirical studies show that FedGreen can substantially reduce the carbon footprints of FL compared to the state-of-the-art while maintaining competitive model accuracy. 
\end{abstract}


\section{Introduction}
With the proliferation of the Internet of Things (IoT) devices and smartphones, Federated learning (FL) has been extensively studied to leverage the valuable data generated by these devices, where the clients do not share data and only model updates are shared and transmitted to the server. 
FL could involve millions of devices to generate an effective global model collaboratively. 
The coordination following FL protocols could cost an unprecedented scale of distributed computing and communication, continuously consuming energy and generating carbon emissions. Meanwhile, such large-scale coordination offers opportunities for optimizing the workload distribution of computation in FL, leading to overall lower carbon emissions.
With over 120 countries committing to the net-zero emissions goal by 2050 \cite{wu2022review}, carbon emissions by cloud and edge computing have become a major concern as data centers and edge servers hosting the FL tasks may still be powered by fossil energy \cite{qiu2021first}. Promoting carbon-intensive computing will jeopardize the net-zero goal.

Existing FL methods rarely considered the carbon emissions of the process or assumed that all participants have uniform carbon intensity levels during the training process, leading to all clients transmitting models of the same size mirroring the global model. However, the real-world scenario presents a stark contrast, where edge devices and cloud computing instances, as FL participants, exhibit varying carbon intensity rates and significant disparities in power sources based on locations and computing architectures. Consequently, it becomes imperative to acknowledge and address this inherent system heterogeneity. When applying this concept to deep neural networks, we must consider the carbon intensity concerning the geographical locations of the clients. This variability can manifest in various ways, such as adjustments in neural network depths (i.e., the number of layers) and widths (i.e., the number of hidden channels). \textit{Can such variability positively contribute to a carbon-aware FL solution for a greener training process?}

\begin{figure}[tb]
\centerline{\includegraphics[scale = 0.6]{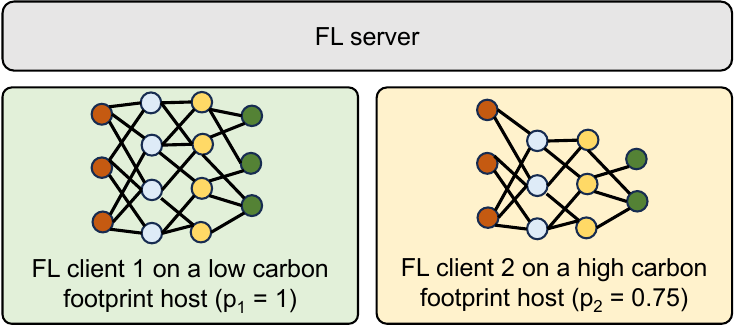}}
\caption{The proposed FedGreen model size adaptation method.  $p_i$ is the remaining rate of neurons. For the clients on hosts with higher carbon footprints, smaller models are sent. For those with lower carbon footprints, larger models are sent. }
\label{fig:intro}
\end{figure}
The model size is a primary indicator of carbon emissions in FL since both the computation and communication cost positively correlate to the model size \cite{qiu2021first}. To adaptively adjust the size of the models in FL, in this paper, we explore ordered dropout to use dynamically configurable sizes of models for local training and model aggregation based on the carbon intensity of the clients \cite{horvath2021fjord}. By employing dropout and sending different model sizes to clients with varying carbon footprints, as shown in Fig. \ref{fig:intro}, we aim to reduce the carbon footprint of the overall process in FL. Additionally, we explore and formulate the adjustment of various parameters, including the local training epochs for FL and the combinations of model size scaling rates, to minimize the overall carbon emissions while maintaining the desired level of accuracy.

The remaining sections are organized as follows. Section \ref{sec:related-work} reviews the related works. In Section \ref{sec:proposed-approach}, we model the carbon emission and introduce our proposed approach. The FedGreen algorithm is then introduced in Section \ref{sec:fedgreen-algo}. We examine our proposed approach with empirical studies in Section \ref{sec:evaluation}. Section \ref{sec:conclusion} concludes the paper.

\section{Related Work}
\label{sec:related-work}
\citet{mcmahan2017communication} proposed FL as a privacy-preserving method for training the model.  In \cite{horvath2021fjord,diao2020heterofl}, the authors proposed methods to send adaptive models for clients with heterogeneous capabilities and training data by using model pruning. The authors in \cite{mei2022resource} used neural composition to address the system heterogeneity and training capacity of the clients. \citet{jiang2022model} proposed a method to reduce the Communication and computation overhead in Federated Learning using model pruning.
\citet{shi2023towards} allowed the clients to participate in cooperative training without directly conducting local gradient calculations. 
\citet{qiu2021first} proposed and evaluated a carbon emission model with sources of energy and consumption in both computation and communication aspects. In \cite{li2023anycostfl}, the authors proposed a method to optimize energy consumption in federated learning using adaptive size models. 
These methods aimed to enhance accuracy and robustness within the context of diverse models in heterogeneous systems. In contrast, FedGreen centers its efforts on reducing the carbon emission in the same rate of convergence accuracy.
FedGreen has also a rigorous study to reduce carbon cost in the context of heterogeneous systems profiles. 
We also provide theoretical analysis for FedGreen.



\section{Proposed Method} \label{sec:proposed-approach}

\subsection{Model Compression}

In this paper, we draw inspiration from previous works. Specifically, we explore ordered dropout (OD) \cite{horvath2021fjord} to organize the representation of knowledge within nested submodels of the original network. The ordered dropout process, initiated by dropping adjacent components of the model rather than random neurons differs from random dropout (RD) \cite{srivastava2014dropout}. It entails resulting in computational advantages and improvements in accuracy rates \cite{horvath2021fjord}. Compared to reducing network depth, reducing network width can be more effective in terms of parameter and memory footprint reduction during inference, a significant advantage for edge devices \cite{yu2018slimmable}.

Our approach involves adapting the network width as the method of choice for resource adjustment. Furthermore, networks with adjusted width still belong to the same model class and share similar fundamental characteristics. As observed in previous studies, this continuity in the model architecture is preferable for maintaining stable training and model aggregation \cite{horvath2021fjord,diao2020heterofl}. Moreover, we utilize a scaling factor denoted as $p$ to adjust the number of active channels in each layer based on the carbon footprints of the computing infrastructure. Slimming down the network width by a factor of $p$ reduces both model size and computational cost by a factor of $p^2$ \cite{horvath2021fjord}. 

To distribute different-sized models to clients based on their locations, we consider a predefined set of representative capacity ratios denoted as $\mathbf{p} \left(\text{e.g.,} \{ p_1, p_2, p_3 \} \right)$. 

\subsection{Optimizing the Carbon Emission}
We observe that the energy usage for each FL client can be categorized into two components: computation usage and communication usage. Suppose there are $N$ clients participating in the FL. We use $c_i \in [1, N]$ to denote the client $i$. A total of $R$ rounds of training will be performed. We use $j \in [1, R]$ to represent the round $j$. We define $C_{FL}$ as the total carbon emission produced in the entire training process of FL. $C_{com}$ denotes the carbon cost in the communication process, and $C_{cmp}$ represents the carbon emission in computation for all $N$ clients over $R$ rounds. Their relationship follows: 
$C_{FL} = {C}_{cmp} + {C_{com}}$.

We apply our problem definition following \cite{wu2022review,li2023anycostfl} to determine the carbon footprints of FL, demonstrating that for each client, its energy usage and carbon emissions have a linear correlation. We use $\theta_{i,j}$ to define the carbon intensity factor, which means the amount of $\text{CO}_2$ produced for each kWh of electricity at the round $j$ for the $c_i$. 

The power consumption (e.g., CPU and memory) of $c_i$ at the round $j$ is denoted by $e_{i,j}^{cmp}$. The energy consumption for client $c_i$ at the  round $j$ is influenced by the duration of the computation, denoted by ${T}^{cmp}_{i,j}$. In Equation (\ref{eq:T-cmp}), $p_{i,j}$ is the scaling factor of the model for the client $c_i$ in the round $j$, and $f_{i,j}$ is the computation frequency of the hardware. $W$ is the total learning time of one data sample. $|\mathcal{D}_i|$ is the number of samples on $c_i$. $E$ is the number of the local rounds. Then we have Equation (\ref{eq:c_cmp}).

\begin{equation}
{T}^{cmp}_{i,j} = \frac{ p_{i,j}^2 E W |\mathcal{D}_i| }{f_{i,j}}
\label{eq:T-cmp}, {C}^{cmp}_{i,j} = {T}^{cmp}_{i,j} e_{i,j}^{cmp} \theta_{i,j}
\end{equation} 

\begin{align}
\begin{split}
C_{cmp} & = \sum_{j=1}^R \sum_{i=1}^N  {C}^{cmp}_{i,j} = \sum_{j=1}^R \sum_{i=1}^N \frac{\theta_{i,j} p_{i,j}^2 E W |\mathcal{D}_i| e_{i,j}^{cmp}  }{f_{i,j}}
\end{split}
\label{eq:c_cmp}
\end{align}

The energy consumption in communication for $N$ clients over $R$ rounds, considering the hardware power for $c_i$ to be $e_{i}^{com}$, can be represented by the formula $C_{com}$. Equation (\ref{eq:cmu}) incorporates several factors, including the size of the model, denoted by $S$, in megabytes, the power of the router, $e_{r}$, and the power of the idle client, $e_{idle}$. The formula also considers the client's upload speed $U$ and download speed $D$, measured in megabytes per second (MBps). In Equation (\ref{eq:T_cmu}), $p_{i,j}$ is the scaling factor of the model. Define the duration of the communication for transmitting model updates for client $c_i$ at the round $j$ as ${T}^{com}_{i,j}$.

\begin{equation}
{T^{com}_{i,j}}= p_{i,j}^2 S \left(\frac{1}{D}+\frac{1}{U}\right)
\label{eq:T_cmu}
\end{equation}

\begin{align}
\begin{split}
    {C}^{com}_{i,j} &= {T}^{com}_{i,j} \theta_{i,j} e_{i,j}^{com} = p_{i,j}^2 S \left(\frac{1}{D}+\frac{1}{U}\right) \theta_{i,j} \left(e_r + e_{i d l e, i} \right)
\end{split}
\end{align}

\begin{align}
\begin{split}
    C_{com} &= \sum_{j=1}^R \sum_{i=1}^N  {C}^{com}_{i,j} \\
    &= \sum_{j=1}^R \sum_{i=1}^N p_{i,j}^2 S\left(\frac{1}{D}+\frac{1}{U}\right) \theta_{i,j} \left(e_r + e_{i d l e, i} \right)
    \label{eq:cmu}
\end{split}
\end{align}

In Equation (\ref{eq:const}), we calculated the total carbon emitted for the $c_i$ in the round $j$. In this context, we use a constant $a_{i,j}$ to represent the constant value associated with the client $c_i$ based on the number of local epochs and other hardware capacity parameters.

\begin{align}
\begin{split}
    {C}^{FL}_{i,j} &= {C}^{cmp}_{i,j} + {C}^{com}_{i,j}\\
    &= p_{i,j}^2\theta_{i,j} \left[ S\left(\frac{1}{D}+\frac{1}{U}\right)  \left(e_r + e_{i d l e, i} \right)+\frac{E W |\mathcal{D}_i| e_{i,j}^{cmp}  }{f_{i,j}} \right] \\
    &=p_{i,j}^2\theta_{i,j}a_{i,j}
    \label{eq:const}
\end{split}
\end{align}

Define a set $\mathbf{p}$ with scaling rates of ${p_1, ..., p_m}$ for a total of $M$ clusters, where $m \in [1, M]$. By tuning $\mathbf{p}$, we can minimize the carbon emissions produced during the training process to reach a target accuracy $\Omega$. Suppose it takes $R$ rounds to reach the target accuracy and if the client $c_i$ belong to cluster $m$, then $p_{i,j} = p_{m}$. In Equation (\ref{eq:total}), $A_{m}$ is the average of $\sum_{i \in N_m} \theta_{i,j} a_{i,j}$ for the cluster $m$ for $R$ rounds, where $N_m$ represents the set of clients in the cluster $m$.

\begin{equation}
A_m = \frac{ \sum_{j=1}^R \sum_{i \in N_m} \theta_{i,j} a_{i,j}}{R}
\end{equation}

\begin{align}
\begin{split}
    C_{FL} &= \sum_{i=1}^N \sum_{j=1}^R {C}^{FL}_{i,j} \\
    &= \sum_{m=1}^M p_{m}^2 R \frac{\sum_{j=1}^R \sum_{i \in N_m} \theta_{i,j} a_{i,j}}{R} = R \sum_{m=1}^M p_{m}^2 A_{m}
\label{eq:total}
\end{split}
\end{align}

\begin{equation}
\operatorname*{argmin}_{p_m \in \mathbf{p}} C_{FL}(\mathbf{p}) = R \sum_{m=1}^M p_{m}^2 A_{m}
\label{eq:arg}
\end{equation}

In Equation (\ref{eq:arg}), we want to minimize the total carbon cost $C_{FL}(\mathbf{p})$ by choosing the $\mathbf{p}$ vector. As we will show in further experiments, $R$ has a positive correlation with the inverse of the mean value of $p_m \in \mathbf{p}$, denoted by $\mu(\mathbf{p})$. $R$ is also positively correlated to the standard deviation of $p_m \in \mathbf{p}$, denoted by $\sigma(p)$. 

\begin{equation}
R \propto \frac{\sigma(\mathbf{p})^\beta}{\mu(\mathbf{p})^\lambda}
\label{eq:mean,sd}
\end{equation}

In Equation (\ref{eq:mean,sd}), $\lambda$ is a constant based on the effect of the mean value on the accuracy rate of the model. $\beta$ is another constant based on the effect of the standard deviation of scaling rate factors on the accuracy rate of the model. These constant parameters are based on the model architecture and diversity of data among the clients. By finding $\mathbf{p}$, which minimizes the Equation (\ref{eq:optimization}) we can find the minimum cost function for the federated system.

\begin{equation}
\operatorname*{argmin}_{p_m \in \mathbf{p}} C_{FL}(\mathbf{p}) = \left(  \frac{\sigma(\mathbf{p})^\beta}{\mu(\mathbf{p})^\lambda} \sum_{m=1}^M p_{m}^2 A_m \right)
\label{eq:optimization}
\end{equation}

\section{The FedGreen Algorithm}
\label{sec:fedgreen-algo}

The FedGreen algorithm aims to identify the set $\mathbf{p}$, i.e., the scaling rate for the model size values that prioritizes minimizing carbon emissions while maintaining reasonable accuracy.
We have observed that carbon intensity of the clients depend on their geolocations and energy sources, which can lead to varying rates of carbon emissions produced by clients in each global round.
In this case, we aim to send models of different sizes to the clients based on the scaling rate of the cluster to which they will be assigned by their carbon profile. Under this method, each cluster of clients would have a scaling rate from parameters in $\mathbf{p}$, which denotes as $p_m$. During the communication phase, the server then broadcasts the compressed model to a set of clients $S_j$ in round $j$. On the client side, each sub-model runs $E$ local iterations using its compressed model. The updated weights $w_{p(i,j)}$ from each device $i$ in round $j$ are then sent to the server. The aggregation process should also be heterogeneous as the client updates are heterogeneous. To achieve this, the following aggregation method, shown in Equation (\ref{eq:Avg}), is utilized to combine the client updates \cite{horvath2021fjord}.

\begin{equation}
\label{eq:Avg}
\boldsymbol{w}_{p_m}^{j+1} \backslash \boldsymbol{w}_{p_{m-1}}^{j+1}=\mathbf{W A}\left(\left\{\boldsymbol{w}_{i_{p_m}}^{(i, j, E)} \backslash \boldsymbol{w}_{p_{m-1}}^{(i, j, E)}\right\}_{i \in \mathcal{S}_j^m}\right)
\end{equation}

In Equation (\ref{eq:Avg}), $\boldsymbol{w}_{p_m}^{j+1} \backslash \boldsymbol{w}_{p_{m-1}}^{j+1}$ are the weights belong to the model of cluster $m$ and not cluster $m-1$ in the round $j+1$. $\mathbf{WA}$ represents the weighted average for updated weights of the clients in ${S}_j^m = \{i \in S_j : p_{i,j} \ge p_m\}$.

\begin{algorithm}
\caption{\textbf{The FedGreen Algorithm}}\label{alg:cap}
\begin{algorithmic}
\State \textbf{Inputs:} Total number of global training rounds $R$. Local training epochs $E$. The whole set of clients $S$. Carbon intensity set for the clients $\theta_{i}$. Number of clusters $M$. Set of scaling rates $\mathbf{p}$. Number of clients N.

\State 1: Initialize \textbf{$w^0$}
\State 2: Server make $M$ clusters of clients based on the $\theta_{i}$
\State 3: Assign specific $p_m$ of  $\mathbf{p}$ to $m$'th cluster.
\State 4: \textbf{for} $j$ $\leftarrow$ 0 to $R-1$ \textbf{do}
\State 5: \quad $S_j \leftarrow$ (random set of $N$ clients)
\State 6: \quad Server sends the weights of $p_i$ submodel to $i$'th client.
\State 7: \quad \textbf{for} $e$ $\leftarrow$ 0 to $E-1$ \textbf{do}
\State 8: \qquad Clients update the weight of the local model.
\State 9: \quad Client $i$ send the updated weights $\boldsymbol{w}^{(i, j, E)}$
\State 10:\quad Server updates \textbf{$w^{j+1}$} as in Equation (\ref{eq:Avg})

\end{algorithmic}
\end{algorithm}

In the proposed FedGreen algorithm, the server performs clustering of the clients based on their $\theta_{i}$ values and number of clusters before starting the global epochs. This clustering helps to group clients with similar carbon intensity factors. Each cluster, denoted by $m$, is assigned a scaling rate of the model compression in $\mathbf{p}$(the clusters with higher average $\theta_{i}$ will usually have lower scaling rates). In each global epoch, the server creates submodels of the original model based on the assigned scaling rate for each cluster. These submodels are then sent to the corresponding clients in their respective clusters. After the clients receive the submodels, they perform local epoch updates by training the submodel with their local data. Once the local weights $\boldsymbol{w}^{(i, j, E)}$ are updated, the clients send the weights back to the server for aggregation. Given the heterogeneity of the models, it follows that the updates for each client's model will also exhibit heterogeneity. To address this variation, we adopt the aggregation rule described in Equation (\ref{eq:Avg}). Finally, the server updates the global model based on the averaged weights received from the clients. This iterative process of clustering, submodel distribution, local updates, and weight averaging allows FedGreen to dynamically adjust the size of the models sent to clients based on the carbon intensity of their location.

\section{Evaluation Results}
\label{sec:evaluation}

\begin{table*}[!htb]

  \centering
  \renewcommand{\arraystretch}{1.2}
  \scalebox{0.89}{\begin{tabular}{ccccccccccc}
    \hline
    \multirow{2}{1cm}{\textbf{Clusters}}&\multirow{2}{1cm}{\textbf{$\mu(\mathbf{p})$}}& \multirow{2}{1.5cm}{\textbf{$\sigma(\mathbf{p})$}}& \multicolumn{2}{c}{$\mathbf{E=1 , \alpha=0.01}$} & \multicolumn{2}{c}{$\mathbf{E=5 , \alpha=0.01}$} & \multicolumn{2}{c}{$\mathbf{E=1 , \alpha=1}$} & \multicolumn{2}{c}{$\mathbf{E=5,\alpha=1}$}\\
    &&&\textbf{rounds} & \textbf{cost} & \textbf{rounds} & \textbf{cost}& \textbf{rounds} & \textbf{cost}& \textbf{rounds} & \textbf{cost}\\
    \hline\hline
    1&1  & 0 & 89 & 18.93 &47 & 74.46& 48& 12.60& 17& 19.65\\ 
    1&0.8  & 0 & 92 & 14.43 & 46& 36.68 &54&10.43& 20& 15.58\\ 
    1&0.6  & 0 & 108 & 11.78 & 51& 22.06 &63&5.89& 23& 9.06 \\ 
    1&0.4  & 0 & 123 & 6.44 & 57 & 13.36 &93&4.45& 34& 7.40\\ 
    1&0.2  & 0 & 200 & 1.98 & 84 & 3.90 &158&1.71& 86& 3.69\\ \hline
    2&0.6  & 0.1 & 115 & 7.81 & 53& 15.11 & 68& 4.53& 25& 8.46\\ 
    2&0.6  & 0.2 & 118 & 6.19 & 57 & 15.30 & 80& 4.17& 27& 8.06 \\ 
    2&0.6  & 0.3 & 142 & 6.83 & 62 & 12.72 & 84& 3.43& 28& 6.29\\ 
    2&0.6  & 0.4 & 148 & 7.69 & 70 & 12.86 & 88& 3.80& 33& 8.18\\ \hline    
    3&0.6  & 0.08 & 110 & 9.40 & 51& 19.50& 65& 5.69& 24& 8.31 \\ 
    3&0.6  & 0.16 & 110 & 8.11 & 54& 17.76  & 65& 4.24& 23& 7.69\\ 
    3&0.6  & 0.24 & 122 & 8.71 &59 & 19.41 & 74& 4.47& 25& 9.72\\ 
    3&0.6  & 0.32 & 123 & 6.79 & 59 & 17.53& 74& 4.36& 30& 8.26 \\ \hline
  \end{tabular}}
  \caption{Effects of $\mu(\mathbf{p})$ and $\sigma(\mathbf{p})$ with varying numbers clusters, local epochs $E = 1, 5$, data heterogeneity $\alpha = 0.01, 1$, and target accuracy levels of 73\%, 78\% on EMNIST.}
  \label{table:res}
\end{table*}
\subsection{Experimental Setups}
\subsubsection{Simulation environment} We assessed the performance of FedGreen using the NVIDIA GeForce 4070 Ti GPU as our hardware platform. In all the scenarios, the idle power consumption $e_{idle} = 10$ Watts. The power consumption of the clients and the router $e_r = 4$ Watts. Also, the average client power consumption rate $e_{client} = 40$ Watts.

\subsubsection{Dataset and model} We utilized the EMNIST \cite{cohen2017emnist}, a dataset for character image classification. It contains 671K $28 \times 28$ images of digits and letters. As for the model, we employed a shallow CNN architecture with nearly 100,000 parameters, consisting of two convolutional layers.

\subsubsection{Configuration}
The total number of clients was set to $N = 80$. In each global round, $10\%$ of the clients, i.e., 8 clients, were selected for sending the model and learning process.
In this experiment, the mini-batch size was 16 for each step, the learning rate was 0.01, and the number of local epochs was set to $E=1$ and $E=5$. We used 80\% of the data for training and 20\% for testing on each client.

\subsubsection{User heterogeneity}
In the context of the EMNIST dataset, we employed a Dirichlet distribution to simulate non-iid data, often denoted as $Dir(\alpha)$, where a smaller $\alpha$ value corresponds to greater data diversity. We illustrated how varying $\alpha$ values impact the statistical diversity within the EMNIST dataset by using 1 and 0.01 for the $\alpha$ values in our experiments.

\subsubsection{Carbon Profile}
We set the carbon intensity rate $\theta$ using two methods. 
\textit{a) Actual carbon intensity. }In the first scenario, we relied on data obtained from six different countries' carbon intensity maps\footnotemark{}\footnotetext{Electricity Maps. https://app.electricitymaps.com/map} whose rates are obtained from governmental sources or on the website Climate Transparency. These countries and their corresponding carbon intensity values in grams per kilowatt-hour (g/kWh) for electricity consumption are as follows: Poland (895 g/kWh), Germany (441 g/kWh), Spain (236 g/kWh), Austria (155 g/kWh), France (47 g/kWh), and Sweden (15 g/kWh). \textit{b) Simulated carbon intensity.} In the second scenario, we employed a simulated carbon intensity approach by considering six different carbon intensity rate values: 0.01 g/kWh, 0.1 g/kWh, 1 g/kWh, 10 g/kWh, 100 g/kWh, and 1000 g/kWh.

\subsection{Experimental Results}

\begin{figure*}[!htb]
  \centering
  \begin{subfigure}{0.24\textwidth}
    \centerline{\includegraphics[scale = 0.3]{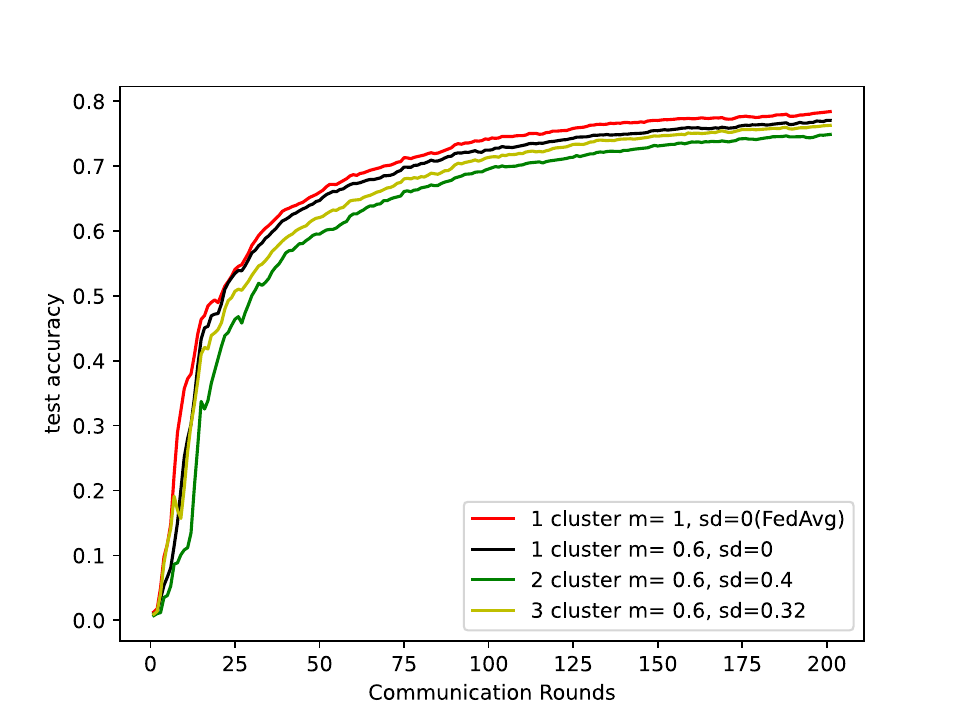}}
    \caption{$E=1 , \alpha=0.01$}
    \label{subfig:a} 
  \end{subfigure}
  \begin{subfigure}{0.24\textwidth}
    \centerline{\includegraphics[scale = 0.3]{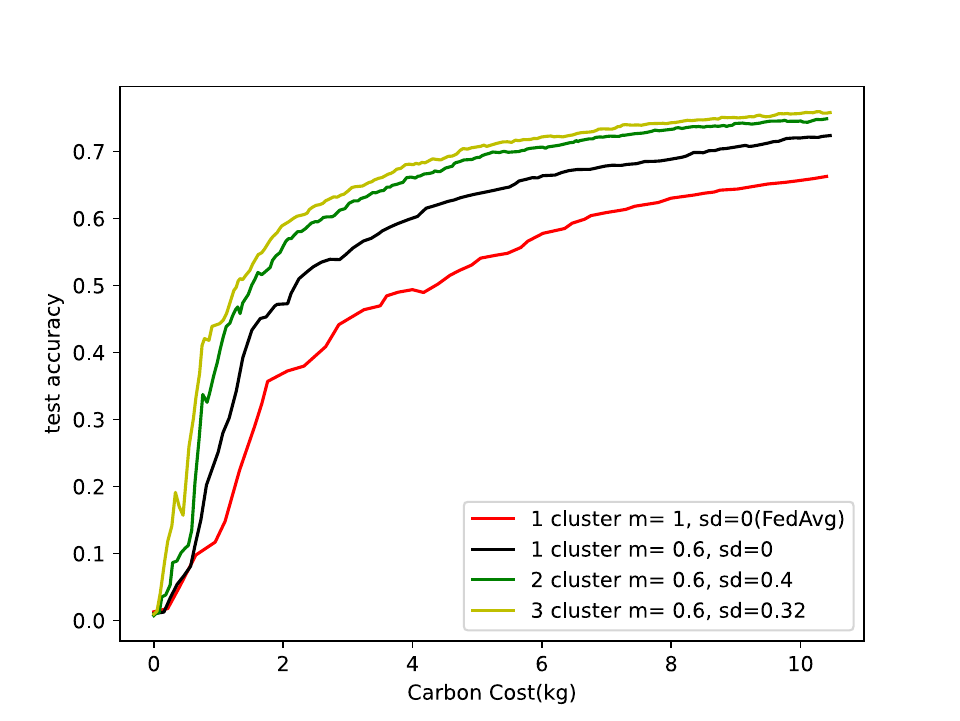}}
    \caption{$E=1 , \alpha=0.01$}
    \label{subfig:b}
  \end{subfigure}
  \begin{subfigure}{0.24\textwidth}
    \centerline{\includegraphics[scale = 0.3]{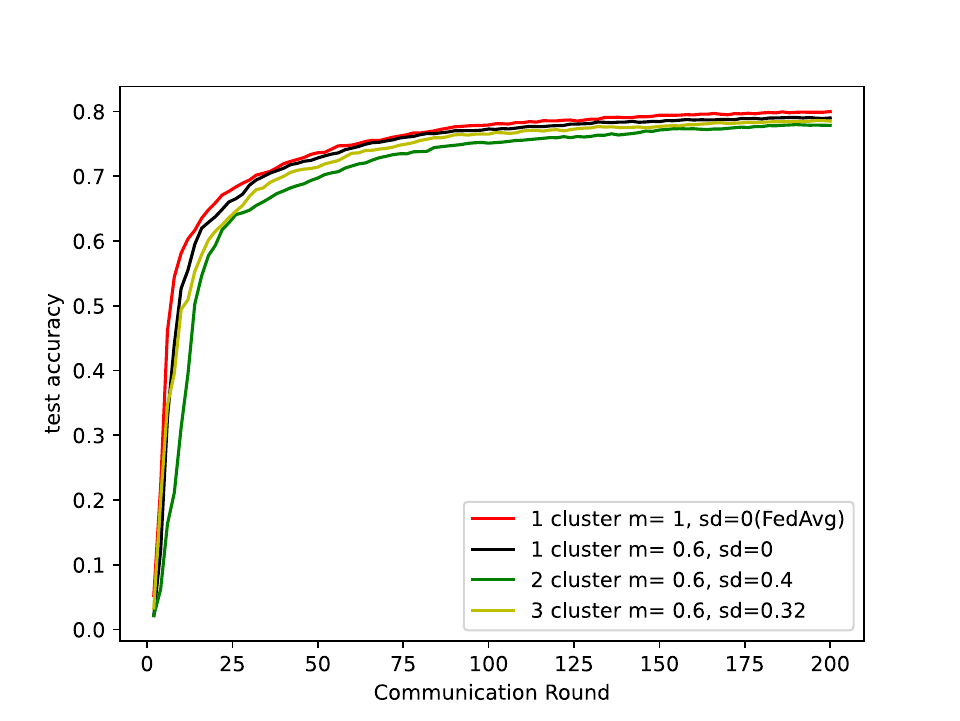}}
       \caption{$E=5 , \alpha=0.01$}
       \label{subfig:c}
  \end{subfigure}
  \begin{subfigure}{0.24\textwidth}
    \centerline{\includegraphics[scale = 0.3]{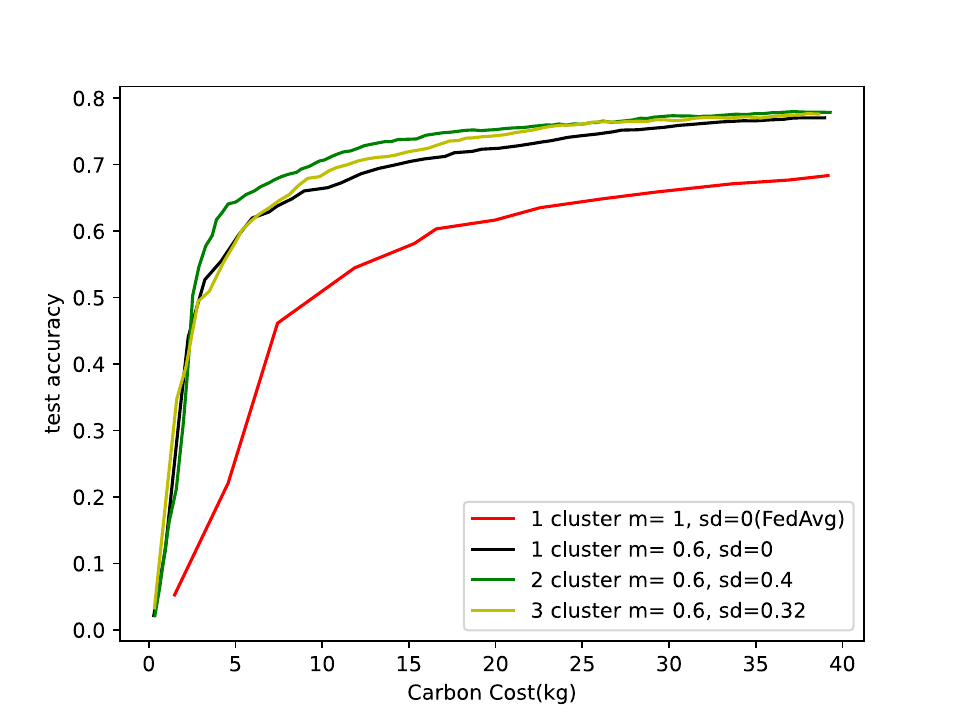}}
    \caption{$E=5 , \alpha=0.01$}
    \label{subfig:d}
  \end{subfigure}

  \begin{subfigure}{0.24\textwidth}
    \centerline{\includegraphics[scale = 0.3]{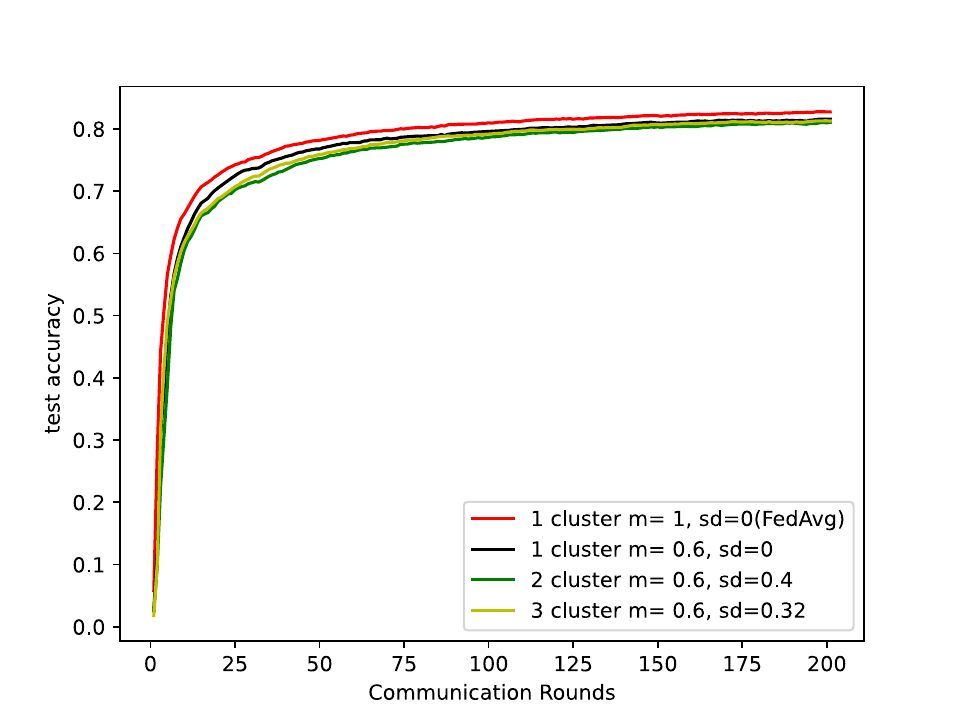}}
    \caption{$E=1 , \alpha=1$}
    \label{subfig:e}
  \end{subfigure}
  \begin{subfigure}{0.24\textwidth}
    \centerline{\includegraphics[scale = 0.3]{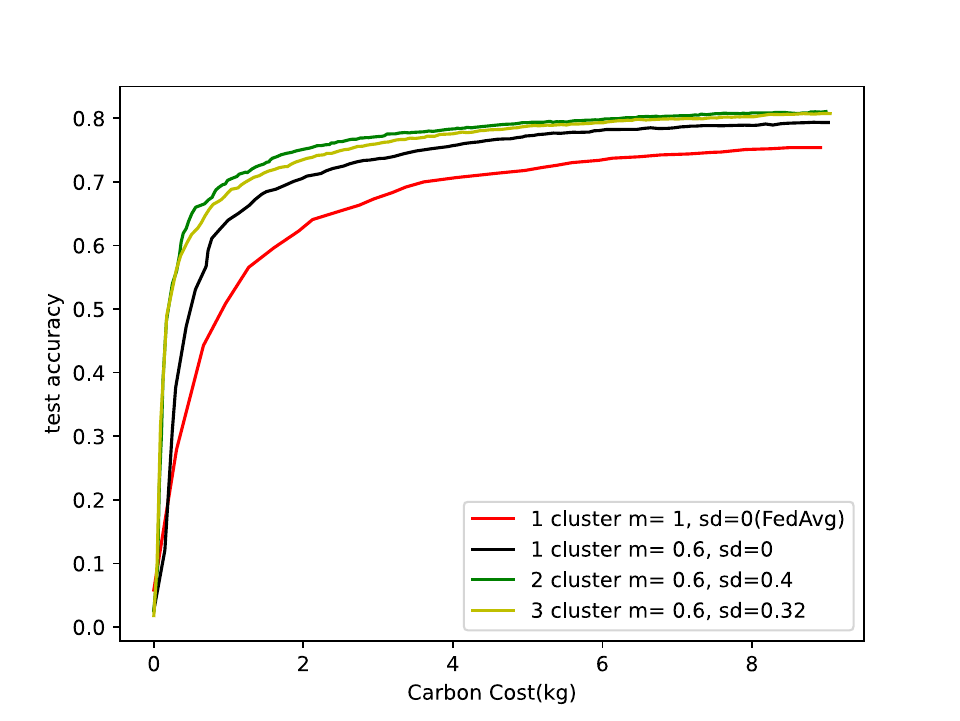}}
    \caption{$E=1 , \alpha=1$}
    \label{subfig:f}
  \end{subfigure}
  \begin{subfigure}{0.24\textwidth}
    \centerline{\includegraphics[scale = 0.3]{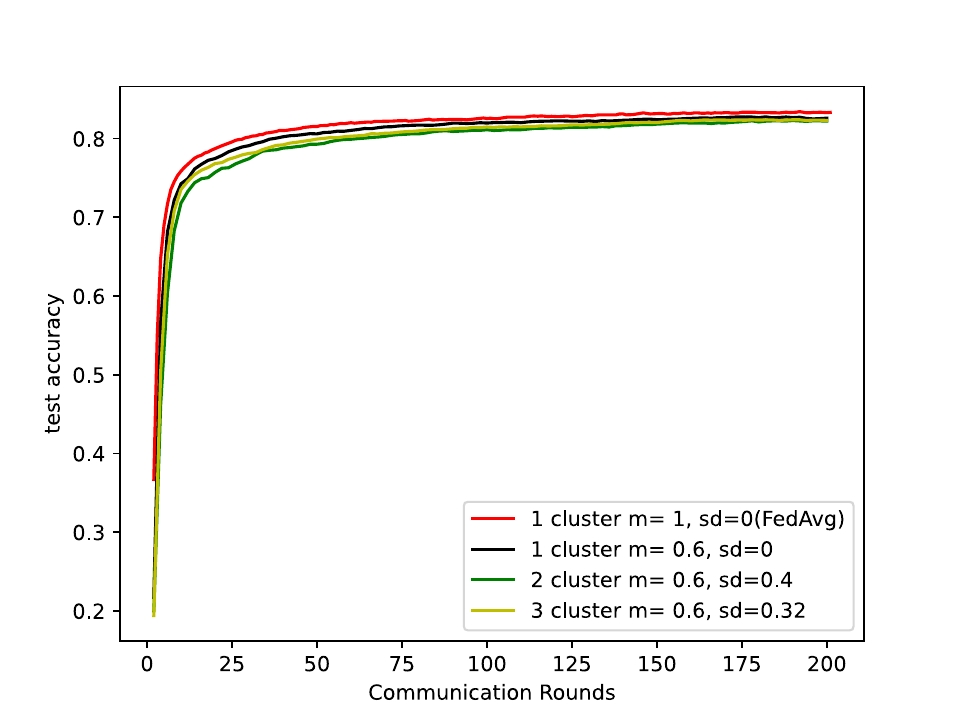}}
       \caption{$E=5 , \alpha=1$}
       \label{subfig:g}
  \end{subfigure}
  \begin{subfigure}{0.24\textwidth}
    \centerline{\includegraphics[scale = 0.3]{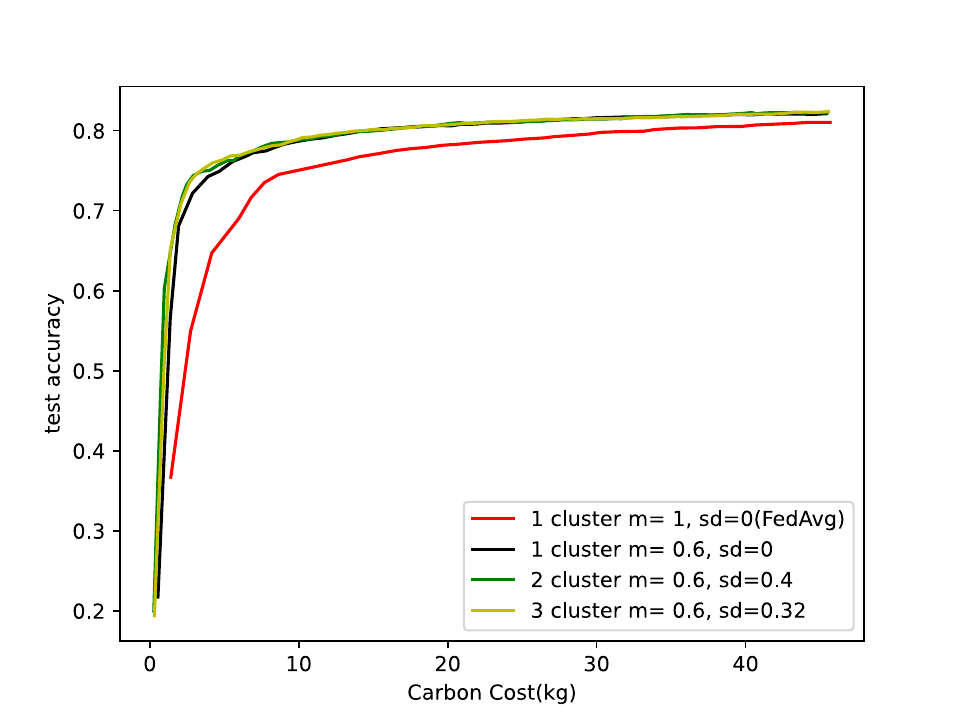}}
    \caption{$E=5 , \alpha=1$}
    \label{subfig:h}
  \end{subfigure}

  \caption{This batch of figures represents a Sensitivity Analysis for Clustering, specifically focusing on varying hyperparameters that exhibit non-identically distributed (non-IID) characteristics and the number of local epochs.}
  \label{fig:two_rows_of_plots}
\end{figure*}

\subsubsection{Mean and standard deviation sensitivity analysis for $\mathbf{p}$}

To gauge the influence of the mean value for $\mathbf{p}$, denoted by $\mu(\mathbf{p})$, we manipulated the scaling rate while keeping the number of clusters at 1.
Specifically, when the $p_i = 1$, the entire model was transmitted to all clients, rendering our algorithm equivalent to the FedAvg \cite{mcmahan2017communication}. Moreover, two distinct scenarios were considered in evaluating the $\mathbf{p}$'s standard deviation assigned to the clusters, denoted by $\sigma(\mathbf{p})$. In these instances, we held $\mu(\mathbf{p})$ while altering the $\sigma(\mathbf{p})$. Notably, with three clusters, a particular value in the $\mathbf{p}$ was fixed to match a consistent $\mu(\mathbf{p})$.
As presented in Table \ref{table:res}, we specifically examined two scenarios: one with $\alpha$ set to 1, indicating a less non-IID data distribution, and another with $\alpha$ set to 0.01, which is a highly non-IID case. Corresponding to different $\alpha$, fixed accuracy levels of 73\% and 78\% were employed.

A reduction in the mean value of the scaling rate is observed to correlate with a decrease in the associated carbon emission cost and a concurrent increase in the number of global epochs. This observed pattern holds across diverse non-IID settings and varying numbers of local epochs. Also, these results prove Equation (\ref{eq:mean,sd}) that the number of rounds has an inverse correlation with the $\mu(\mathbf{p})$ which the power of this relation $\lambda$  is based on the training and model parameters.

\begin{figure*}[tb]
  \centering
  \begin{subfigure}{0.24\textwidth}
    \centerline{\includegraphics[scale = 0.3]{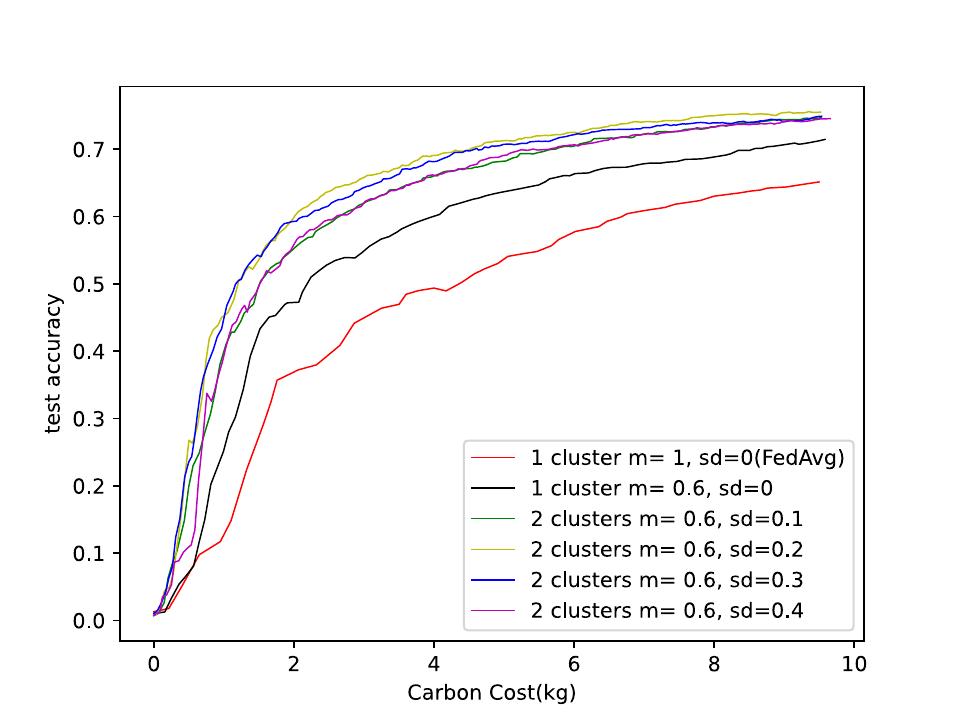}}
    \caption{Real intensity profile}
    \label{subfig:a2} 
  \end{subfigure}
  \begin{subfigure}{0.24\textwidth}
    \centerline{\includegraphics[scale = 0.3]{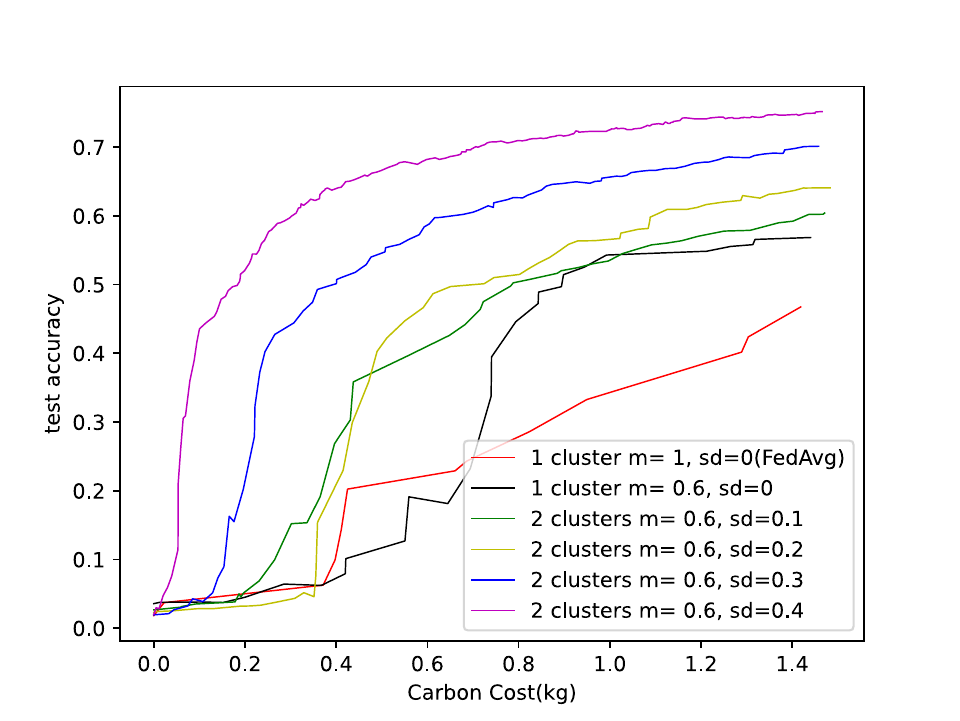}}
    \caption{Simulated intensity profile}
    \label{subfig:b2}
  \end{subfigure}
    \begin{subfigure}{0.24\textwidth}
    \centerline{\includegraphics[scale = 0.3]{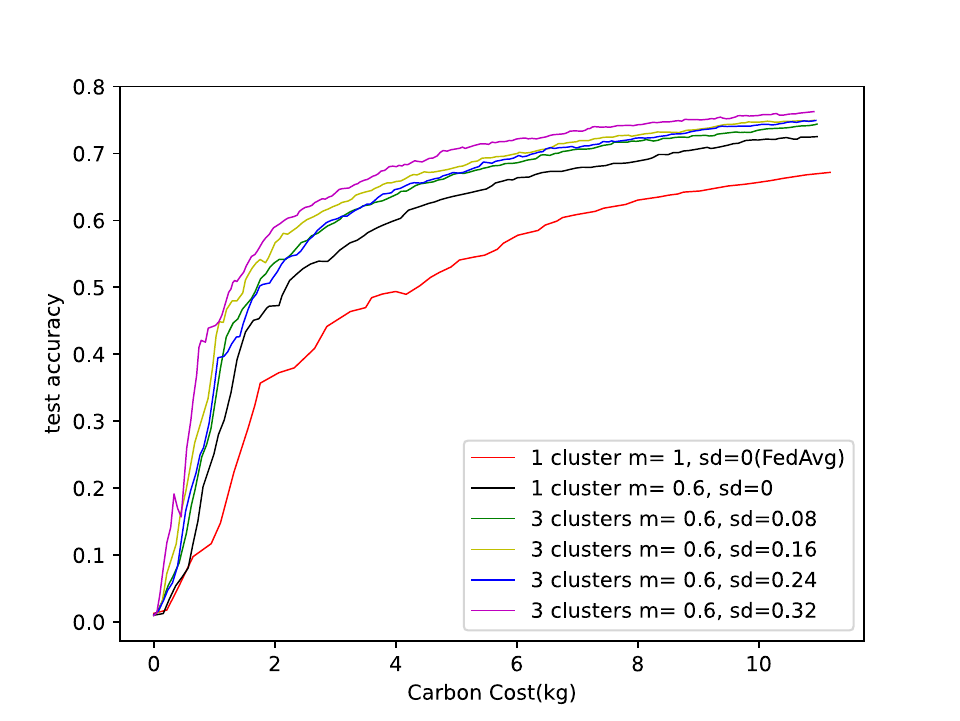}}
    \caption{Real intensity profile}
    \label{subfig:c2} 
  \end{subfigure}
  \begin{subfigure}{0.24\textwidth}
    \centerline{\includegraphics[scale = 0.3]{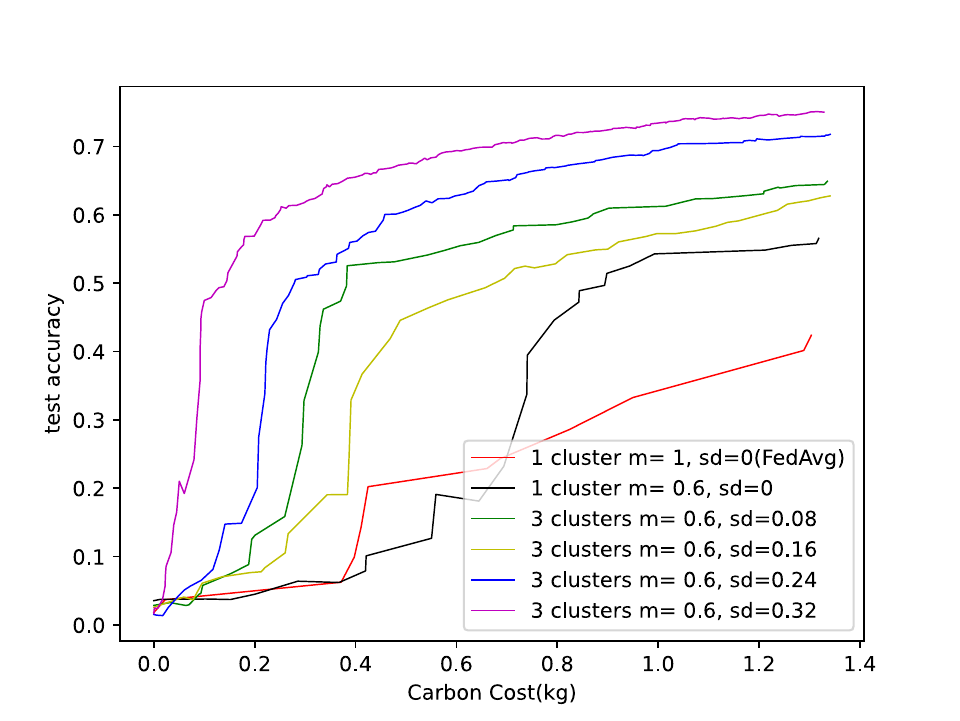}}
    \caption{Simulated intensity profile}
    \label{subfig:d2}
  \end{subfigure}
  \caption{ This batch of figures represents a sensitivity analysis for carbon intensity profile using two different approaches(Simulated and Real) for the carbon profile of the clients in two and three clusters scenarios.}
  \label{fig:row_of_plots}
\end{figure*}

Our investigation into standard deviation analysis involved the examination of two distinct scenarios. In the initial scenario, we considered two clusters with a constant mean value for their scaling rate. Subsequently, in the second scenario, we explored three clusters, wherein we maintained the scaling rate of one cluster at the mean value of the three clusters while varying the scaling rates of the remaining two clusters. The results demonstrate that an increase in standard deviation in both the 2-cluster and 3-cluster cases leads to a consistent elevation in the number of epochs required to achieve a predetermined accuracy across all experiments. This escalation is particularly pronounced when the data exhibits greater non-IID characteristics. Specifically, when we had 2 clusters with 1 epoch for each global round, the impact on the increase in epochs was more pronounced, highlighting the sensitivity of the system's performance to non-IID data distribution. These results prove Equation (\ref{eq:mean,sd}) which the number of rounds correlates with the $\sigma(\mathbf{p})$ which the power of this relation $\beta$ is based on the data diversity between clients.The observed trend indicates that as the standard deviation rises, the number of epochs required for convergence increases while the carbon emissions per global epoch decreases. This underscores a discernible trade-off associated with increasing the standard deviation, suggesting the need for a judicious balance to achieve optimal outcomes.

From Table \ref{table:res}, it is observed that $\sigma(\mathbf{p})$ is more sensitive when $\alpha$ is smaller. Consequently, in instances of high data heterogeneity, a preliminary warm-up step in determining hyperparameters becomes imperative.

\subsubsection{Clustering Sensitivity Analysis} 

We delved into the varying cluster numbers and compared outcomes across various scenarios with baseline approaches.
We introduced an augmented degree of standard deviation for scenarios involving two and three clusters, where the vector $\mathbf{p}$ assumed the values of ${0.2, 1}$ and ${0.2, 0.6, 1}$, respectively, in concordance with the number of clusters under consideration. In this experiment, we explored four distinct scenarios characterized by variations in the $\alpha$ parameter, set at values of 1 and 0.01.
Additionally, we manipulated the local epoch count, considering both $E=1$ and $E=5$. In Fig. \ref{fig:two_rows_of_plots}, each of the four distinct parameter configurations exhibits two distinct types of graphical representation. One depicts test accuracy as a function of the number of communication rounds, while the other illustrates test accuracy in relation to the carbon emissions measured in kilograms (kg). In these experiments, we used the real intensity scenario for the carbon profile of the clients. 

In Fig. \ref{subfig:b} and Fig. \ref{subfig:d}, it is evident that a substantial accuracy gap($9.6\%$ and $9.4\%$) exists between the baseline, represented by FedAvg, and the scenarios involving two and three clusters. This discrepancy is noticeably significant, indicating the method's competence in handling highly non-IID data. Furthermore, $2.5\%$ and $3.4\%$ accuracy difference is also apparent between scenarios featuring two and three clusters and the one-cluster setting with equivalent mean values, particularly when employing only a single epoch. The observed gap diminishes as the number of epochs increases. This reduction in the gap can be attributed to the lower difference in the convergence rate of models with varying sizes, particularly when more epochs are employed in the process, as evident from Fig. \ref{subfig:c}.

For a more IID data distribution in Fig. \ref{subfig:e}, the disparity in test accuracy related to communication rounds between the two-cluster, three-cluster, and baselines is notably smaller. However, in Fig. \ref{subfig:f}, a discernible gap still persists, particularly in comparison to the FedAvg baseline, emphasizing the influence of the clustering configurations on carbon emission. In the scenario characterized by more IID data and a local epoch count of 5, the gap between the different cluster configurations is minimized. However, there still exists a noticeable accuracy gap between these configurations and the FedAvg baseline, as illustrated in Fig. \ref{subfig:h}.

\subsubsection{Carbon Intensity Profile Analysis} 
In this experiment, we conducted a comparison between the real carbon intensity profile, which includes data from six European countries, and the simulated carbon intensity profile. The simulated profile is characterized by a more significant value disparity between green and non-green clients, specifically concerning carbon intensity data. To assess the impact of this contrast, we introduced changes in the standard deviation of the $\mathbf{p}$ vector for scenarios involving two and three clusters, as demonstrated in Fig. \ref{fig:row_of_plots}. As the plots illustrate, when the carbon profiles become more diverged, the gap in test accuracy between the three-cluster and two-cluster scenarios widens. In such cases, a higher standard deviation exerts a more pronounced influence, and the optimal point appears to align with this parameter.

\section{Conclusion}
\label{sec:conclusion} 
In this paper, we proposed FedGreen to mitigate carbon emissions in FL by dynamically assigning model sizes to clients based on the carbon intensity rate of their respective locations. Additionally, we aimed to choose parameters optimally for our model size of different clusters of clients to achieve the best possible trade-off between accuracy rate and carbon emissions. Through a series of experiments, we demonstrated our proposed method's significant advantages and effectiveness.

\bibliography{main.bbl}
\end{document}